\documentclass{article}

\usepackage{iclr2020_conference,times}
\usepackage{hyperref}
\usepackage{url}

\usepackage{microtype}
\usepackage{graphicx}
\usepackage{subfigure}
\usepackage{booktabs} 

\usepackage{amsmath} 
\usepackage{graphicx} 
\usepackage{placeins} 
\usepackage{xfrac} 

\usepackage{hyperref}

\RequirePackage{algorithm}
\RequirePackage{algorithmic}

\newcommand{\tb}{\textbf} 

\title{Learned Step Size Quantization}

\author{Steven K. Esser \thanks{corresponding author sesser@us.ibm.com},
Jeffrey L. McKinstry,
Deepika Bablani, \\
\textbf{Rathinakumar Appuswamy,
Dharmendra S. Modha}\\
\\
IBM Research \\
San Jose, California, USA \\
}

\iclrfinalcopy 
\begin{document}

\maketitle

\begin{abstract}
Deep networks run with low precision operations at inference time offer power and space advantages over high precision alternatives, but need to overcome the challenge of maintaining high accuracy as precision decreases.
Here, we present a method for training such networks, Learned Step Size Quantization, that achieves the highest accuracy to date on the ImageNet dataset when using models, from a variety of architectures, with weights and activations quantized to 2-, 3- or 4-bits of precision, and that can train 3-bit models that reach full precision baseline accuracy.
Our approach builds upon existing methods for learning weights in quantized networks by improving how the quantizer itself is configured.
Specifically, we introduce a novel means to estimate and scale the task loss gradient at each weight and activation layer's quantizer step size, such that it can be learned in conjunction with other network parameters.
This approach works using different levels of precision as needed for a given system and requires only a simple modification of existing training code.

\end{abstract}

\section{Introduction}

Deep networks are emerging as components of a number of revolutionary technologies, including image recognition \citep{krizhevsky2012imagenet}, speech recognition \citep{hinton2012deep}, and driving assistance \citep{xu2017end}.
Unlocking the full promise of such applications requires a system perspective where task performance, throughput, energy-efficiency, and compactness are all critical considerations to be optimized through co-design of algorithms and deployment hardware.
Current research seeks to develop methods for creating deep networks that maintain high accuracy while reducing the precision needed to represent their activations and weights, thereby reducing the computation and memory required for their implementation.
The advantages of using such algorithms to create networks for low precision hardware has been demonstrated in several deployed systems \citep{Esser11441, jouppi2017datacenter, qiu2016going}.

It has been shown that low precision networks can be trained with stochastic gradient descent by updating high precision weights that are quantized, along with activations, for the forward and backward pass \citep{courbariaux2015binaryconnect, Esser11441}.
This quantization is defined by a mapping of real numbers to the set of discrete values supported by a given low precision representation (often integers with 8-bits or less).
We would like a mapping for each quantized layer that maximizes task performance,
but it remains an open question how to optimally achieve this.

To date, most approaches for training low precision networks have employed uniform quantizers, which can be configured by a single step size parameter (the width of a quantization bin), though more complex nonuniform mappings have been considered \citep{polino2018model}.
Early work with low precision deep networks used a simple fixed configuration for the quantizer \citep{hubara2016binarized,Esser11441},
while starting with \citet{rastegari2016xnor}, later work focused on fitting the quantizer to the data, either based on statistics of the data distribution \citep{li2016ternary,zhou2016dorefa,cai2017deep,mckinstry2018discovering} or seeking to minimize quantization error during training \citep{choi2018learning,zhang2018lq}.
Most recently, work has focused on using backpropagation with stochastic gradient descent to learn a quantizer that minimizes task loss \citep{DBLP:journals/corr/ZhuHMD16,mishra2017apprentice,choi2018pact,choi2018bridging,jung2018joint,baskin2018nice,polino2018model}.

While attractive for their simplicity, fixed mapping schemes based on user settings place no guarantees on optimizing network performance,
and quantization error minimization schemes might perfectly minimize quantization error and yet still be non optimal if a different quantization mapping actually minimizes task error.
Learning the quantization mapping by seeking to minimize task loss is appealing to us as it directly seeks to improve on the metric of interest. 
However, as the quantizer itself is discontinuous, such an approach requires approximating its gradient, which existing methods have done in a relatively coarse manner that ignore the impact of transitions between quantized states \citep{choi2018pact,choi2018bridging,jung2018joint}.

\begin{table*}[bt]
	\caption{Comparison of low precision networks on ImageNet.
	Techniques compared are QIL \citep{jung2018joint}, FAQ \citep{mckinstry2018discovering}, LQ-Nets \citep{zhang2018lq}, PACT \citep{choi2018pact},  Regularization \citep{choi2018learning}, and NICE \citep{baskin2018nice}.
	}
	\label{table:top1}
	\setlength\tabcolsep{6pt}
	\begin{center}
	\small
	\begin{tabular}{l l cccc c cccc}
		\toprule
		& & \multicolumn{4}{c}{Top-1 Accuracy @ Precision} & & \multicolumn{4}{c}{Top-5 Accuracy @ Precision} \\
		Network & Method & 2 & 3 & 4 & 8 & & 2 & 3 & 4 & 8 \\
		\midrule
		ResNet-18 	& & \multicolumn{4}{c}{{\textit{Full precision: 70.5}}} & & \multicolumn{4}{c}{{\textit{Full precision: 89.6}}} \\
					&	LSQ (Ours) 	 	& \tb{67.6}	& \tb{70.2}& \tb{71.1}& \tb{71.1}& 	& \tb{87.6}	& \tb{89.4}	& \tb{90.0}	& \tb{90.1} \\
					& 	QIL 				& 65.7 	& 69.2 	& 70.1 	& 		& 	&  \\
					&	FAQ				&		& 		& 69.8	& 70.0	& 	& 		&		& 89.1	& 89.3 \\			
					& 	LQ-Nets 			& 64.9	& 68.2 	& 69.3 	&  		&	& 85.9 	& 87.9	& 88.8	& \\		
					&	PACT 			& 64.4	& 68.1 	& 69.2 	&  		&	& 85.6	& 88.2	& 89.0	& \\
					&	NICE			& 		& 67.7	& 69.8	&		&	& 		& 87.9	& 89.21	& \\			
					& 	Regularization		& 61.7	& 		& 67.3	& 68.1	&	& 84.4	&		& 87.9	& 88.2 \\	
		\midrule
		ResNet-34	& & \multicolumn{4}{c}{{\textit{Full precision: 74.1}}} & & \multicolumn{4}{c}{{\textit{Full precision: 91.8}}} \\
					&	LSQ (Ours) 	 	& \tb{71.6}& \tb{73.4} & \tb{74.1}& \tb{74.1}& 	& \tb{90.3}	& \tb{91.4}	& \tb{91.7}	&\tb{91.8} \\
					& 	QIL 				& 70.6 	& 73.1 	& 73.7 	& 		& 	&  \\					
					&	LQ-Nets			& 69.8	& 71.9	&		&		& 	& 89.1	& 90.2	&		& \\
					&	NICE			& 		& 71.7	& 73.5	&		&	& 		& 90.8	& 91.4	& \\
					&	FAQ				& 		& 		& 73.3	& 73.7	&	&		&		& 91.3	& 91.6 \\	
		\midrule		
		ResNet-50	& & \multicolumn{4}{c}{{\textit{Full precision: 76.9}}} & & \multicolumn{4}{c}{{\textit{Full precision: 93.4}}} \\		
					&	LSQ (Ours) 	 	& \tb{73.7}&\tb{75.8}	& \tb{76.7}	& \tb{76.8}	&	& \tb{91.5}	& \tb{92.7}	& 93.2	& \tb{93.4} \\
					&	PACT			& 72.2	& 75.3	& 76.5	&		& 	& 90.5	& 92.6	& 93.2	& \\
					& 	NICE			& 		& 75.1   	& 76.5	&		&	& 		& 92.3	& \tb{93.3}	& \\		
					&	FAQ				& 		&		& 76.3	& 76.5 	&	&		&		& 92.9	& 93.1 \\	
					& 	LQ-Nets			& 71.5	& 74.2	& 75.1	&		&	& 90.3	& 91.6	& 92.4	& \\	
		\midrule		
		ResNet-101	& & \multicolumn{4}{c}{{\textit{Full precision: 78.2}}} & & \multicolumn{4}{c}{{\textit{Full precision: 94.1}}} \\
					&	LSQ (Ours) 	 	& \tb{76.1}	& \tb{77.5}	& \tb{78.3}	& \tb{78.1}	&	& \tb{92.8}	& \tb{93.6}	& \tb{94.0}	& \tb{94.0} \\		
		\midrule		
		ResNet-152	& & \multicolumn{4}{c}{{\textit{Full precision: 78.9}}} & & \multicolumn{4}{c}{{\textit{Full precision: 94.3}}} \\		
					&	LSQ (Ours) 		& \tb{76.9}	& \tb{78.2}	& \tb{78.5}	& \tb{78.5}	& 	& \tb{93.2}	& \tb{93.9}	& \tb{94.1} & \tb{94.2} \\
					&	FAQ				& 		& 		& 78.4	& \tb{78.5}	& 	& 		& 		& \tb{94.1} 	& 94.1 \\
		\midrule		
		VGG-16bn	& & \multicolumn{4}{c}{{\textit{Full precision: 73.4}}} & & \multicolumn{4}{c}{{\textit{Full precision: 91.5}}} \\		
					&	LSQ (Ours) 		& \tb{71.4}	& \tb{73.4}	& \tb{74.0}	& 73.5 &	& \tb{90.4}	& \tb{91.5}	& \tb{92.0}	& \tb{91.6} \\	
					&	FAQ				& 		& 		& 73.9	& \tb{73.7} &	& 	& 		& 91.7	& \tb{91.6} \\
		\midrule		
		Squeeze		& & \multicolumn{4}{c}{{\textit{Full precision: 67.3}}} & & \multicolumn{4}{c}{{\textit{Full precision: 87.8}}} \\		
		Next-23-2x	&	LSQ (Ours) 		& \tb{53.3}	& \tb{63.7}	&\tb{ 67.4}	& \tb{67.0} &	& \tb{77.5}	& \tb{85.4}	& \tb{87.8}	& \tb{87.7} \\
		\bottomrule		
	\end{tabular}
	\end{center}
\end{table*}

Here, we introduce a new way to learn the quantization mapping for each layer in a deep network, \textit{Learned Step Size Quantization} (LSQ), that improves on prior efforts with two key contributions.
First, we provide a simple way to approximate the gradient to the quantizer step size that is sensitive to quantized state transitions, 
arguably providing for finer grained optimization when learning the step size as a model parameter.
Second, we propose a simple heuristic to bring the magnitude of step size updates into better balance with weight updates, which we show improves convergence.
The overall approach is usable for quantizing both activations and weights, and works with existing methods for backpropagation and stochastic gradient descent.
Using LSQ to train several network architectures on the ImageNet dataset, we demonstrate 
significantly better accuracy than prior quantization approaches (Table \ref{table:top1}) and,
for the first time that we are aware of, demonstrate the milestone of 3-bit quantized networks reaching full precision network accuracy (Table \ref{table:kd}).

\section{Methods}

We consider deep networks that operate at inference time using low precision integer operations for computations in convolution and fully connected layers, requiring quantization of the weights and activations these layers operate on.
Given data to quantize $v$, quantizer step size $s$, the number of positive and negative quantization levels $Q_P$ and $Q_N$, respectively, we define a quantizer that computes $\bar{v}$, a quantized and integer scaled representation of the data, and $\hat{v}$, a quantized representation of the data at the same scale as $v$:
\begin{equation}
	\label{eq:quantization1}
	\bar{v} = \lfloor clip( \sfrac{v}{s}, -Q_N, Q_P) \rceil,
\end{equation}
\begin{equation}
	\label{eq:quantization2}
	\hat{v} = \bar{v} \times s.
\end{equation}
Here, $clip(z,r_1,r_2)$ returns $z$ with values below $r_1$ set to $r_1$ and values above $r_2$ set to $r_2$, and $\lfloor z \rceil$ rounds $z$ to the nearest integer.
Given an encoding with $b$ bits, for unsigned data (activations) $Q_N=0$ and $Q_P=2^b-1$ and for signed data (weights) $Q_N=2^{b-1}$ and $Q_P=2^{b-1}-1$.

\begin{figure}[ht]
	\begin{center}
  	\centerline{\includegraphics[width=3.5in]{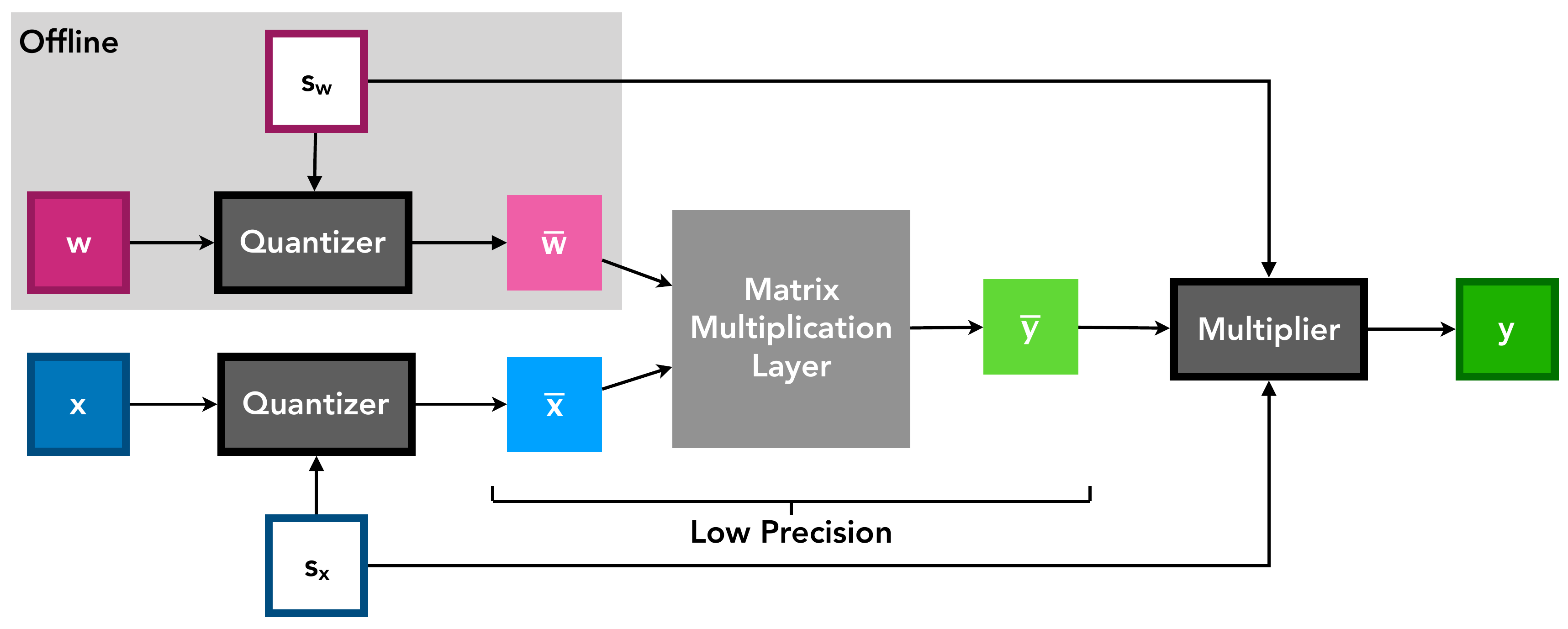}}
  	\caption{
  	Computation of a low precision convolution or fully connected layer, as envisioned here.
  	}
  	\label{fig:quantization}
	\end{center}
\end{figure}

For inference, $\bar{w}$ and $\bar{x}$ values can be used as input to low precision integer matrix multiplication units underlying convolution or fully connected layers, and the output of such layers then rescaled by the step size using a relatively low cost high precision scalar-tensor multiplication, a step that can potentially be algebraically merged with other operations such as batch normalization (Figure \ref{fig:quantization}).

\subsection{Step Size Gradient}
\label{methods:LSQ}

LSQ provides a means to learn $s$ based on the training loss by introducing the following gradient through the quantizer to the step size parameter:
\begin{equation}\label{eq:lsq}
\frac{\partial{\hat{v}}}{\partial{s}} =
\begin{cases}
-\sfrac{v}{s} + \lfloor \sfrac{v}{s} \rceil			& \text{if $-Q_N < \sfrac{v}{s} < Q_P$} \\
-Q_N									& \text{if $\sfrac{v}{s} \le -Q_N$} \\
Q_P									& \text{if $\sfrac{v}{s} \ge Q_P$} \\
\end{cases}
\end{equation}
This gradient is derived by using the straight through estimator \citep{bengio2013estimating} to approximate the gradient through the round function as a pass through operation (though leaving the round itself in place for the purposes of differentiating down stream operations), and differentiating all other operations in Equations \ref{eq:quantization1} and \ref{eq:quantization2} normally.

\begin{figure}[ht]
	\begin{center}
  	\centerline{\includegraphics[width=\columnwidth]{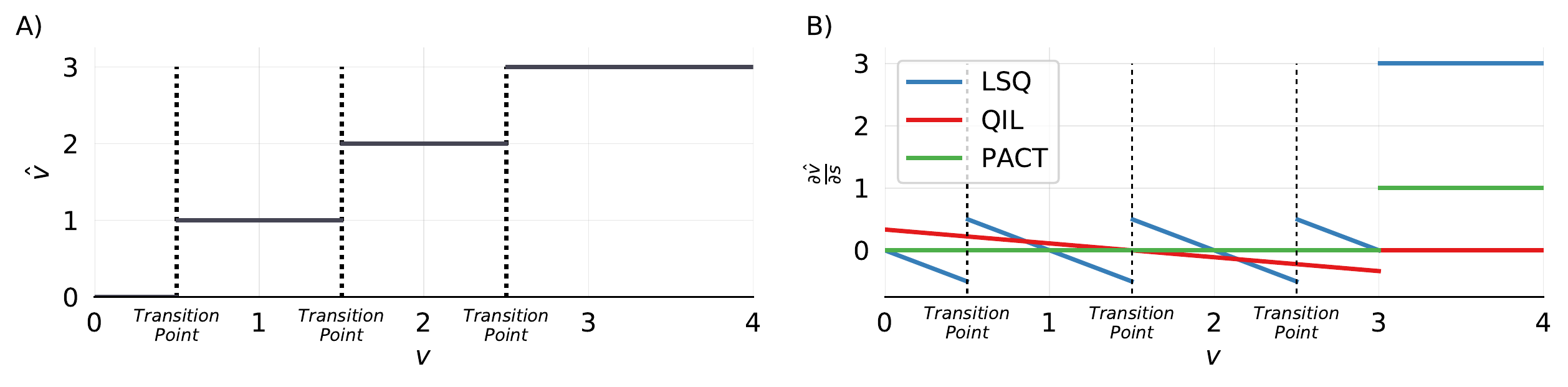}}
	\caption{Given $s=1$, $Q_N=0$, $Q_P=3$, A) quantizer output and B)
gradients of the quantizer output with respect to step size, $s$, for LSQ, or a related parameter controlling the width of the quantized domain (equal to $s(Q_P+Q_N)$) for QIL \citep{jung2018joint} and PACT \citep{choi2018pact}.
The gradient employed by LSQ is sensitive to the distance between $v$ and each transition point,
whereas the gradient employed by QIL \citep{jung2018joint} is sensitive only to the distance from quantizer clip points,
and the gradient employed by PACT \citep{choi2018pact} is zero everywhere below the clip point.
Here, we demonstrate that networks trained with the LSQ gradient reach higher accuracy than those trained with the QIL or PACT gradients in prior work.
	}
	\label{fig:gradients}
  	\end{center}
\end{figure}

This gradient differs from related approximations (Figure \ref{fig:gradients}),
which instead either learn a transformation of the data that occurs completely prior to the discretization itself \citep{jung2018joint}, or estimate the gradient by removing the round operation from the forward equation, algebraically canceling terms, and then differentiating such that $\sfrac{\partial{\hat{v}}}{\partial{s}}=0$ where $-Q_N < \sfrac{v}{s} < Q_P$ \citep{choi2018pact,choi2018bridging}.
In both such previous approaches, the relative proximity of $v$ to the transition point between quantized states does not impact the gradient to the quantization parameters.
However, one can reason that the closer a given $v$ is to a quantization transition point, the more likely it is to change its quantization bin ($\bar{v}$) as a result of a learned update to $s$ (since a smaller change in $s$ is required), thereby resulting in a large jump in $\hat{v}$.  Thus, we would expect $\sfrac{\partial{\hat{v}}}{\partial{s}}$ to increase as the distance from $v$ to a transition point decreases, and indeed we observe this relationship in the LSQ gradient.  It is appealing that this gradient naturally falls out of our simple quantizer formulation and use of the straight through estimator for the round function.

For this work, each layer of weights and each layer of activations has a distinct step size, represented as an fp32 value, initialized to $\sfrac{2\langle|v|\rangle}{\sqrt{Q_P}}$, computed on either the initial weights values or the first batch of activations, respectively.

\subsection{Step Size Gradient Scale}
\label{methods:gradscale}

It has been shown that good convergence is achieved during training where the ratio of average update magnitude to average parameter magnitude is approximately the same for all weight layers in a network \citep{you2017large}.
Once learning rate has been properly set, this helps to ensure that all updates are neither so large as to lead to repeated overshooting of local minima, nor so small as to lead to unnecessarily long convergence time. 
Extending this reasoning, we consider that each step size should also have its update magnitude to parameter magnitude proportioned similarly to that of weights.
Thus, for a network trained on some loss function $L$, the ratio
\begin{equation}\label{eq:R}
R = 
\left.
\frac{
	\nabla_s L
}{
	s
}
\right/
\frac{
	\lVert \nabla_w L \rVert
}{
	\lVert w \rVert
}
\end{equation}
should on average be near 1, where $\lVert z \rVert$ denotes the $l_2$-norm of $z$. 
However, we expect the step size parameter to be smaller as precision increases (because the data is quantized more finely), and step size updates to be larger as the number of quantized items increases (because more items are summed across when computing its gradient).
To correct for this, we multiply the step size loss by a gradient scale, $g$, where for weight step size $g=\sfrac{1}{\sqrt{N_WQ_P}}$ and for activation step size $g=\sfrac{1}{\sqrt{N_FQ_P}}$, where $N_W$ is the number of weights in a layer and $N_f$ is the number of features in a layer.
In section \ref{sec:gradscaleres} we demonstrate that this improves trained accuracy, and we provide reasoning behind the specific scales chosen in the Section \ref{a:gradscale} of the Appendix.

\subsection{Training}
\label{methods:training}

Model quantizers are trained with LSQ by making their step sizes learnable parameters with loss gradient computed using the quantizer gradient described above, while other model parameters can be trained using existing techniques.
Here, we employ a common means of training quantized networks \citep{courbariaux2015binaryconnect},
where full precision weights are stored and updated, quantized weights and activations are used for forward and backward passes, the gradient through the quantizer round function is computed using the straight through estimator \citep{bengio2013estimating} such that
\begin{equation}\label{eq:xgrad}
	\frac{\partial \hat{v}}{\partial v} = 
	\begin{cases}
		1 & \text{if $-Q_N < \sfrac{v}{s} < Q_P$} \\
		0 & \text{otherwise},
	\end{cases}
\end{equation}
and stochastic gradient descent is used to update parameters.

For simplicity during training, we use $\hat{v}$ as input to matrix multiplication layers, which is algebraically equivalent to the previously described inference operations.
We set input activations and weights to either 2-, 3-, 4-, or 8-bit for all matrix multiplication layers except the first and last, which always use 8-bit, as making the first and last layers high precision has become standard practice for quantized networks and demonstrated to provide a large benefit to performance.  All other parameters are represented using fp32.
All quantized networks are initialized using weights from a trained full precision model with equivalent architecture before fine-tuning in the quantized space, which is known to improve performance \citep{sung2015resiliency,zhou2016dorefa,mishra2017apprentice,mckinstry2018discovering}.

Networks were trained with a momentum of 0.9, using a softmax cross entropy loss function, and cosine learning rate decay without restarts \citep{DBLP:journals/corr/LoshchilovH16a}.
Under the assumption that the optimal solution for 8-bit networks is close to the full precision solution \citep{mckinstry2018discovering}, 8-bit networks were trained for 1 epoch while all other networks were trained for 90 epochs.
The initial learning rate was set to 0.1 for full precision networks, 0.01 for 2-, 3-, and 4-bit networks and to 0.001 for 8-bit networks.
All experiments were conducted on the ImageNet dataset \citep{russakovsky2015imagenet}, using pre-activation ResNet \citep{he2016deep}, VGG \citep{simonyan2014very} with batch norm, or SqueezeNext \citep{gholami2018squeezenext}.
All full precision networks were trained from scratch, except for VGG-16bn, for which we used the pretrained version available in the PyTorch model zoo.
Images were resized to 256 $\times$ 256, then a 224 $\times$ 224 crop was selected for training, with horizontal mirroring applied half the time.  At test time, a 224 $\times$ 224 centered crop was chosen.
We implemented and tested LSQ in PyTorch.

\section{Results}

\subsection{Weight Decay}

We expect that reducing model precision will reduce a model's tendency to overfit, and thus also reduce the regularization in the form of weight decay necessary to achieve good performance.
To investigate this, we performed a hyperparameter sweep on weight decay for ResNet-18 (Table  \ref{table:weightdecay}),
and indeed found that lower precision networks reached higher accuracy with less weight decay.
Performance was improved by reducing weight decay by half for the 3-bit network, and reducing it by a quarter for the 2-bit network. We used these weight decay values for all further experiments.

\begin{table}[h]
	\caption{ResNet-18 top-1 accuracy for various weight decay values.}
	\label{table:weightdecay}
	\setlength\tabcolsep{6pt}
	\begin{center}
	\begin{small}
	\begin{tabular}{r c c c c}
		Weight Decay & 2-bit & 3-bit & 4-bit & 8-bit \\
		\midrule
		$10^{-4}$ 				& $66.9$ 			& $70.1$ 			& $\mathbf{71.0}$	& $\mathbf{71.1}$ \\
		$0.5 \times 10^{-4}$ 		& $67.3$ 			& $\mathbf{70.2}$ 	& $70.9$			& $71.1$ \\
		$0.25 \times 10^{-4}$ 	& $\mathbf{67.6}$ 	& $70.0$ 			& $70.9$ 			& $71.0$ \\
		$0.125 \times 10^{-4}$ 	& $67.4$ 			& $66.9$ 			& $70.8$ 			& $71.0$ \\
		\bottomrule
	\end{tabular}
	\end{small}
	\end{center}
\end{table}

\subsection{Comparison with Other Approaches}

We trained several networks using LSQ and compare accuracy with other quantized networks and full precision baselines (Table \ref{table:top1}).
To facilitate comparison, we only consider published models that quantize all convolution and fully connected layer weights and input activations to the specified precision, except for the first and last layers which may use higher precision (as for the LSQ models).
In some cases, we report slightly higher accuracy on full precision networks than in their original publications, which we attribute to our use of cosine learning rate decay \citep{DBLP:journals/corr/LoshchilovH16a}.

We found that LSQ achieved a higher top-1 accuracy than all previous reported approaches for 2-, 3- and 4- bit networks with the architectures considered here.
For nearly all cases, LSQ also achieved the best-to-date top-5 accuracy on these networks, and best-to-date accuracy on 8-bit versions of these networks.
In most cases, we found no accuracy advantage to increasing precision from 4-bit to 8-bit.
It is worth noting that the next best low precision method \citep{jung2018joint} used progressive fine tuning (sequentially training a full precision to 5-bit model, then the 5-bit model to a 4-bit model, and so on), significantly increasing training time and complexity over our approach which fine tunes directly from a full precision model to the precision of interest.

It is interesting to note that when comparing a full precision to a 2-bit precision model, top-1 accuracy drops only $2.9$ for ResNet-18, but $14.0$ for SqueezeNext-23-2x.
One interpretation of this is that the SqueezeNext architecture was designed to maximize performance using as few parameters as possible, which may have placed it at a design point extremely sensitive to reductions in precision.

\subsection{Accuracy vs. Model Size}

\begin{figure}[ht]
	\begin{center}
  	\centerline{\includegraphics[width=5in]{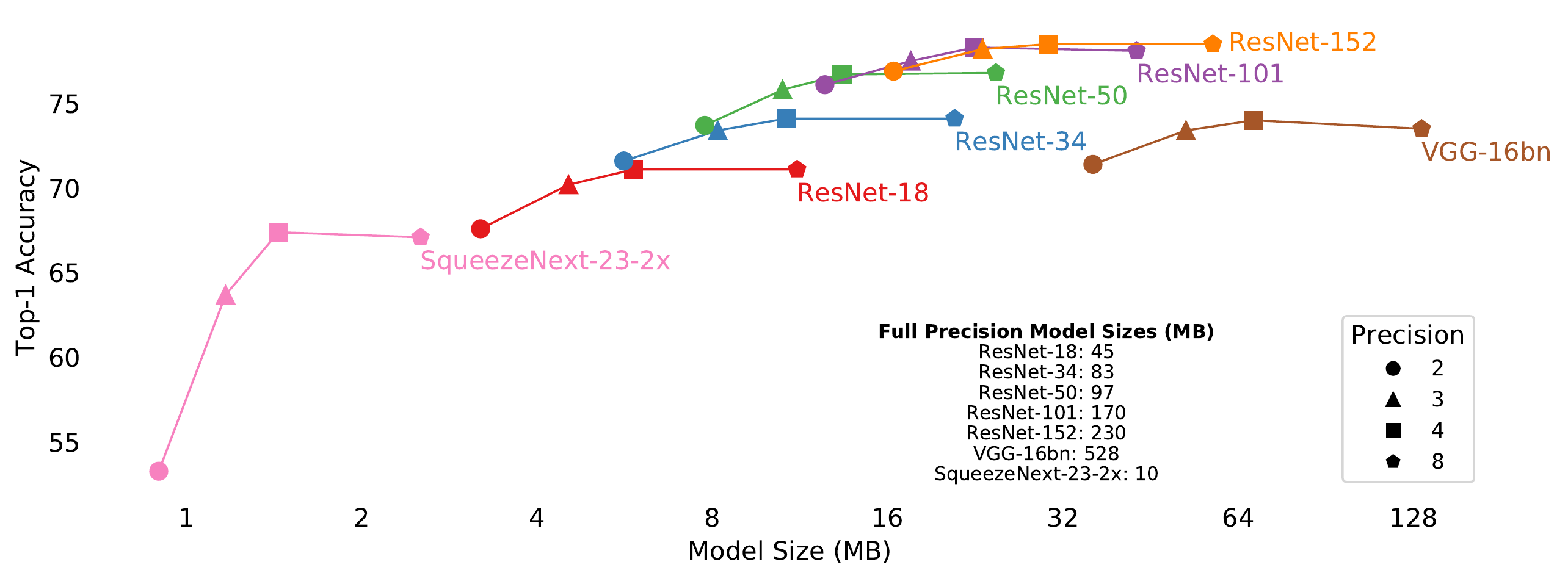}}
  \caption{
  	Accuracy vs. model size for the networks considered here show some 2-bit networks provide the highest accuracy at a given model size.  Full precision model sizes are inset for reference.
}
    	\label{fig:acc_v_size}
  	\end{center}
\end{figure}

For a model size limited application, it is important to choose the highest performing model that fits within available memory limitations.  To facilitate this choice, we plot here network accuracy against corresponding model size (Figure \ref{fig:acc_v_size}).

We can consider the frontier of best performance for a given model size of the architectures considered here.
On this metric, we can see that 2-bit ResNet-34 and ResNet-50 networks offer an absolute advantage over using a smaller network, but with higher precision.
We can also note that at all precisions, VGG-16bn exists below this frontier, which is not surprising as this network was developed prior to a number of recent innovations in achieving higher performance with fewer parameters.

\subsection{Step Size Gradient Scale Impact}
\label{sec:gradscaleres}

\begin{figure}[ht]
	\begin{center}
  	\centerline{\includegraphics[width=3in]{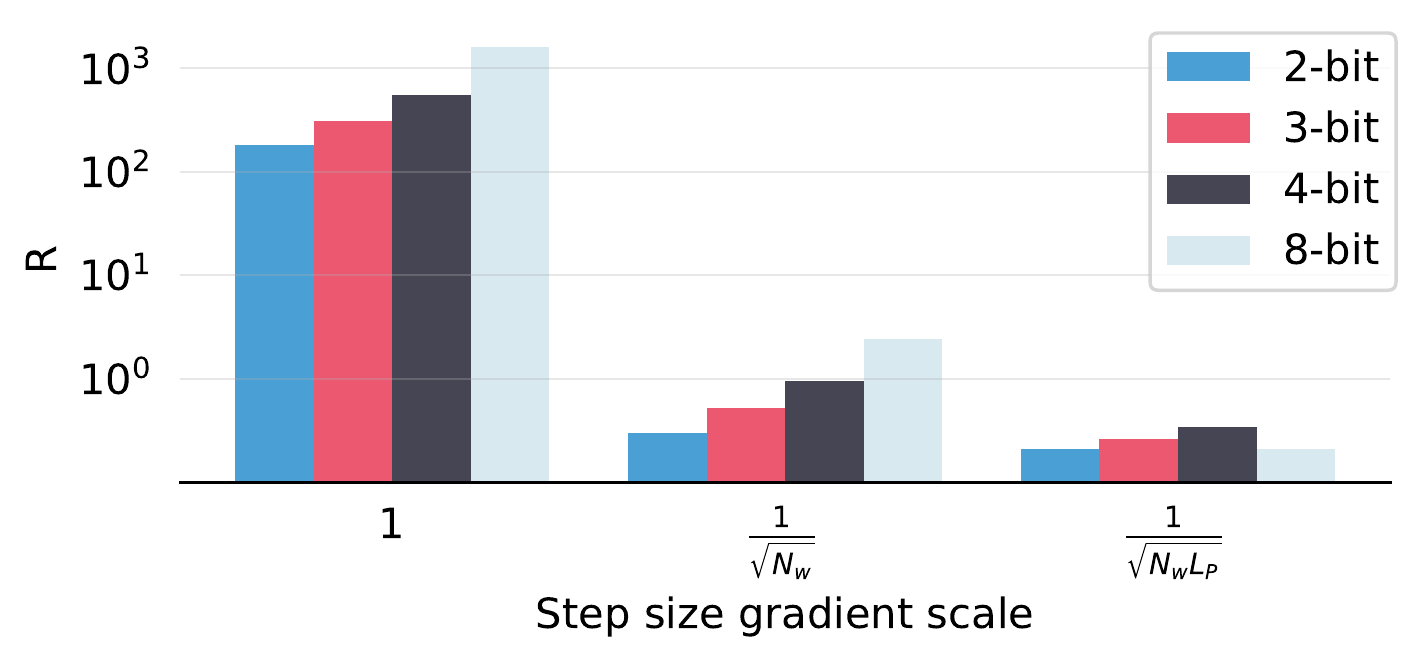}}
  \caption{
	Relative parameter update magnitudes given different step size gradient scales.  A gradient scale of $\sfrac{1}{N_WQ_P}$ better balances relative step size and weight gradient magnitudes (right vs. left).
}
    	\label{fig:results_grad}
  	\end{center}
\end{figure}

To demonstrate the impact of the step size gradient scale (Section \ref{methods:gradscale}), we measured $R$ (see Equation \ref{eq:R}) averaged 
across $500$ iterations in the middle of the first training epoch for ResNet-18,
using different step size gradient scales (the network itself was trained with the scaling as described in the methods to avoid convergence problems).
With no scaling, we found that relative to parameter size, updates to step size were $2$ to $3$ orders of magnitude larger than updates to weights, and this imbalance increased with precision, with the 8-bit network showing almost an order of magnitude greater imbalance than the 2-bit network (Figure \ref{fig:results_grad}, left). 
Adjusting for the number of weights per layer ($g=\sfrac{1}{\sqrt{N_W}}$), the imbalance between step size and weights largely went away, through the imbalance across precision remained (Figure \ref{fig:results_grad}, center).
Adjusting for the number of number of weights per layer and precision ($g=\sfrac{1}{\sqrt{N_WQ_P}}$), this precision dependent imbalance was largely removed as well (Figure \ref{fig:results_grad}, right).
 
We considered network accuracy after training a 2-bit ResNet-18 using different step size gradient scales (Table \ref{table:gscale}).
Using the network with the full gradient scale ($g=\sfrac{1}{\sqrt{N_WQ_P}}$ and $g=\sfrac{1}{\sqrt{N_FQ_P}}$ for weight and activation step size respectively) as baseline, we found that adjusting only for weight and feature count led to a $0.3$ decrease in top-1 accuracy, and when no gradient scale was applied the network did not converge unless we dropped the initial learning rate.
Dropping the initial learning rate in multiples of ten, the best top-1 accuracy we achieved using no gradient scale was $3.4$ below baseline, using an initial learning rate of $0.0001$.
Finally, we found that using the full gradient scaling with an additional ten-fold increase or decrease also reduced top-1 accuracy.
Overall, this suggests a benefit to our chosen heuristic for scaling the step size loss gradient.

\begin{table}[h]
	\caption{Top-1 accuracy for various gradient scale values and learning rates for 2-bit ResNet-18.}
	\label{table:gscale}
	\setlength\tabcolsep{6pt}
	\begin{center}
	\begin{small}
	\begin{tabular}{l c c}
    		\toprule
		Gradient scale 		& Learning Rate 	& Accuracy \\
		\midrule
		$\bf{\sfrac{1}{\sqrt{NQ_P}}}$	& \bf{0.01}		& $\bf{67.6}$	\\ 
		\midrule		
		$\sfrac{1}{\sqrt{N}}$		& 0.01		& $67.3$		\\
		$ 1 $				& 0.01		&  Did not converge \\
		$ 1 $				& 0.0001		&  64.2 \\
		\midrule		 	
		$\sfrac{10}{\sqrt{NQ_P}}$		& 0.01		& $67.4$		\\ 	
		$\sfrac{1}{10\sqrt{NQ_P}}$ 	& 0.01		& $67.3$		\\
		\bottomrule
	\end{tabular}
	\end{small}
	\end{center}
\end{table}

\subsection{Cosine Learning Rate Decay Impact}

We chose to use cosine learning rate decay in our experiments as it removes the need to select learning rate schedule hyperparameters, is available in most training frameworks, and does not increase training time.
To facilitate comparison with results in other publications that use step-based learning rate decay, we trained a 2-bit ResNet-18 model with LSQ for 90 epochs, using an initial learning rate of $0.01$, which was multiplied by $0.1$ every $20$ epochs.
This model reached a top-1 accuracy of $67.2$, a reduction of $0.4$ from the equivalent model trained with cosine learning rate decay, but still marking an improvement of $1.5$ over the next best training method (see Table \ref{table:top1}).

\subsection{Quantization Error}
\label{sec:qe}

We next sought to understand whether LSQ learns a solution that minimizes quantization error (the distance between $\hat{v}$ and $v$ on some metric), despite such an objective not being explicitly encouraged.
For this purpose, for a given layer we define the final step size learned by LSQ as $\hat{s}$ and let $S$ be the set of discrete values $\{0.01\hat{s}, 0.02\hat{s}, ..., 20.00\hat{s}\}$.
For each layer, on a single batch of test data we computed the value of $s \in S$ that minimizes
mean absolute error, $\langle |(\hat{v}(s) - v)| \rangle$,
mean square error, $\langle (\hat{v}(s) - v)^2 \rangle$,
and Kullback-Leibler divergence, $\int p(v) \log p(v) - \int p(v) \log q(\hat{v}(s))$ where $p$ and $q$ are probability distributions.
For purposes of relative comparison, we ignore the first term of Kullback-Leibler divergence, as it does not depend on $\hat{v}$, and approximate the second term as $ - \text{E}[ \log(q(\hat{v}(s)))]$, where the expectation is over the sample distribution.

For a 2-bit ResNet-18 model we found $\hat{s}=0.949 \pm 0.206$ for activations and $\hat{s}=0.025 \pm 0.019$ for weights (mean $\pm$ standard deviation).
The percent absolute difference between $\hat{s}$ and the value of $s$ that minimizes quantization error, averaged across activation layers was $50\%$ for mean absolute error, $63\%$ for mean square error, and $64\%$ for Kullback-Leibler divergence,
and averaged across weight layers, was $47\%$ for  mean absolute error, $28\%$ for mean square error, and $46\%$ for Kullback-Leibler divergence.
This indicates that LSQ learns a solution that does not in fact minimize quantization error.
As LSQ achieves better accuracy than approaches that directly seek to minimize quantization error, this suggests that simply fitting a quantizer to its corresponding data distribution may not be optimal for task performance.

\subsection{Improvement with knowledge-distillation}

To better understand how well low precision networks can reproduce full precision accuracy, we combined LSQ with same-architecture knowledge distillation, which has been shown to improve low precision network training \citep{mishra2017apprentice}.
Specifically, we used the distillation loss function of \citet{hinton2015distilling} with temperature of 1 and equal weight given to the standard loss and the distillation loss (we found this gave comparable results to weighting the the distillation loss two times more or less than the standard loss on 2-bit ResNet-18).  The teacher network was a trained full precision model with frozen weights and of the same architecture as the low precision network trained.
As shown in Table \ref{table:kd}, this improved performance, with top-1 accuracy increasing by up to $1.1$ (3-bit ResNet-50), and with 3-bit networks reaching the score of the full precision baseline (see Table \ref{table:top1} for comparison).
As a control, we also used this approach to distill from the full precision teacher to a full precision (initially untrained) student with the same architecture, which did not lead to an improvement in the student network accuracy beyond training the student alone.
These results reinforce previous work showing that knowledge-distillation can help low precision networks catch up to full precision performance \citep{mishra2017apprentice}.

\begin{table}[h]
	\caption{Accuracy for low precision networks trained with LSQ and knowledge distillation, which is improved over using LSQ alone, with 3-bit networks reaching the accuracy of full precision (32-bit) baselines (shown for comparison).}
	\label{table:kd}
	\setlength\tabcolsep{6pt}
	\begin{center}
	\small
	\begin{tabular}{l ccccc c ccccc}
		\toprule
		& \multicolumn{5}{c}{Top-1 Accuracy @ Precision} & & \multicolumn{5}{c}{Top-5 Accuracy @ Precision} \\
		Network  & 2 & 3 & 4 & 8 & \textit{32} & & 2 & 3 & 4 & 8 & \textit{32}\\
		\midrule
		ResNet-18 	& 67.9	& 70.6	& 71.2	& 71.1	& \textit{70.5}	& 	& 88.1	& 89.7	& 90.1	& 90.1	& \textit{89.6} \\
		ResNet-34 	& 72.4	& 74.3	& 74.8	& 74.1	& \textit{74.1}	&	& 90.8	& 91.8	& 92.1	& 91.7 	& \textit{91.8} \\		
		ResNet-50 	& 74.6	& 76.9	& 77.6	& 76.8	& \textit{76.9}	& 	& 92.1	& 93.4	& 93.7	& 93.3 	& \textit{93.4} \\				
		\bottomrule
	\end{tabular}
	\end{center}
\end{table}

\section{Conclusions}

The results presented here demonstrate that on the ImageNet dataset across several network architectures, LSQ exceeds the performance of all prior approaches for creating quantized networks.
We found best performance when rescaling the quantizer step size loss gradient based on layer size and precision.
Interestingly, LSQ does not appear to minimize quantization error, whether measured using mean square error, mean absolute error, or Kullback-Leibler divergence.
The approach itself is simple, requiring only a single additional parameter per weight or activation layer.

Although our goal is to train low precision networks to achieve accuracy equal to their full precision counterparts, it is not yet clear whether this goal is achievable for 2-bit networks, which here reached accuracy several percent below their full precision counterparts.  However, we found that such 2-bit solutions for state-of-the-art networks are useful in that they can give the best accuracy for the given model size, for example, with an 8MB model size limit, a 2-bit ResNet-50 was better than a 4-bit ResNet-34 (Figure \ref{fig:acc_v_size}).

This work is a continuation of a trend towards steadily reducing the number of bits of precision necessary to achieve good performance across a range of network architectures on ImageNet.
While it is unclear how far it can be taken, it is noteworthy that the trend towards higher performance at lower precision strengthens the analogy between artificial neural networks and biological neural networks, which themselves employ synapses represented by perhaps a few bits of information \citep{bartol2015nanoconnectomic} and single bit spikes that may be employed in small spatial and/or temporal ensembles to provide low bit width data representation.
Analogies aside, reducing network precision while maintaining high accuracy is a promising means of reducing model size and increasing throughput to provide performance advantages in real world deployed deep networks.

\FloatBarrier


\bibliography{refs}

\begin{thebibliography}{33}
\providecommand{\natexlab}[1]{#1}
\providecommand{\url}[1]{\texttt{#1}}
\expandafter\ifx\csname urlstyle\endcsname\relax
  \providecommand{\doi}[1]{doi: #1}\else
  \providecommand{\doi}{doi: \begingroup \urlstyle{rm}\Url}\fi

\bibitem[Bartol~Jr et~al.(2015)Bartol~Jr, Bromer, Kinney, Chirillo, Bourne,
  Harris, and Sejnowski]{bartol2015nanoconnectomic}
Thomas~M Bartol~Jr, Cailey Bromer, Justin Kinney, Michael~A Chirillo,
  Jennifer~N Bourne, Kristen~M Harris, and Terrence~J Sejnowski.
\newblock Nanoconnectomic upper bound on the variability of synaptic
  plasticity.
\newblock \emph{Elife}, 4:\penalty0 e10778, 2015.

\bibitem[Baskin et~al.(2018)Baskin, Liss, Chai, Zheltonozhskii, Schwartz,
  Girayes, Mendelson, and Bronstein]{baskin2018nice}
Chaim Baskin, Natan Liss, Yoav Chai, Evgenii Zheltonozhskii, Eli Schwartz, Raja
  Girayes, Avi Mendelson, and Alexander~M Bronstein.
\newblock Nice: Noise injection and clamping estimation for neural network
  quantization.
\newblock \emph{arXiv preprint arXiv:1810.00162}, 2018.

\bibitem[Bengio et~al.(2013)Bengio, L{\'e}onard, and
  Courville]{bengio2013estimating}
Yoshua Bengio, Nicholas L{\'e}onard, and Aaron Courville.
\newblock Estimating or propagating gradients through stochastic neurons for
  conditional computation.
\newblock \emph{arXiv preprint arXiv:1308.3432}, 2013.

\bibitem[Cai et~al.(2017)Cai, He, Sun, and Vasconcelos]{cai2017deep}
Zhaowei Cai, Xiaodong He, Jian Sun, and Nuno Vasconcelos.
\newblock Deep learning with low precision by half-wave gaussian quantization.
\newblock In \emph{Proceedings of the IEEE Conference on Computer Vision and
  Pattern Recognition}, pp.\  5918--5926, 2017.

\bibitem[Choi et~al.(2018{\natexlab{a}})Choi, Chuang, Wang, Venkataramani,
  Srinivasan, and Gopalakrishnan]{choi2018bridging}
Jungwook Choi, Pierce I-Jen Chuang, Zhuo Wang, Swagath Venkataramani,
  Vijayalakshmi Srinivasan, and Kailash Gopalakrishnan.
\newblock Bridging the accuracy gap for 2-bit quantized neural networks (qnn).
\newblock \emph{arXiv preprint arXiv:1807.06964}, 2018{\natexlab{a}}.

\bibitem[Choi et~al.(2018{\natexlab{b}})Choi, Wang, Venkataramani, Chuang,
  Srinivasan, and Gopalakrishnan]{choi2018pact}
Jungwook Choi, Zhuo Wang, Swagath Venkataramani, Pierce I-Jen Chuang,
  Vijayalakshmi Srinivasan, and Kailash Gopalakrishnan.
\newblock Pact: Parameterized clipping activation for quantized neural
  networks.
\newblock \emph{arXiv preprint arXiv:1805.06085}, 2018{\natexlab{b}}.

\bibitem[Choi et~al.(2018{\natexlab{c}})Choi, El-Khamy, and
  Lee]{choi2018learning}
Yoojin Choi, Mostafa El-Khamy, and Jungwon Lee.
\newblock Learning low precision deep neural networks through regularization.
\newblock \emph{arXiv preprint arXiv:1809.00095}, 2018{\natexlab{c}}.

\bibitem[Courbariaux et~al.(2015)Courbariaux, Bengio, and
  David]{courbariaux2015binaryconnect}
Matthieu Courbariaux, Yoshua Bengio, and Jean-Pierre David.
\newblock Binaryconnect: Training deep neural networks with binary weights
  during propagations.
\newblock In \emph{Advances in neural information processing systems}, pp.\
  3123--3131, 2015.

\bibitem[Esser et~al.(2016)Esser, Merolla, Arthur, Cassidy, Appuswamy,
  Andreopoulos, Berg, McKinstry, Melano, Barch, di~Nolfo, Datta, Amir, Taba,
  Flickner, and Modha]{Esser11441}
Steven~K. Esser, Paul~A. Merolla, John~V. Arthur, Andrew~S. Cassidy,
  Rathinakumar Appuswamy, Alexander Andreopoulos, David~J. Berg, Jeffrey~L.
  McKinstry, Timothy Melano, Davis~R. Barch, Carmelo di~Nolfo, Pallab Datta,
  Arnon Amir, Brian Taba, Myron~D. Flickner, and Dharmendra~S. Modha.
\newblock Convolutional networks for fast, energy-efficient neuromorphic
  computing.
\newblock \emph{Proceedings of the National Academy of Sciences}, 113\penalty0
  (41):\penalty0 11441--11446, 2016.

\bibitem[Gholami et~al.(2018)Gholami, Kwon, Wu, Tai, Yue, Jin, Zhao, and
  Keutzer]{gholami2018squeezenext}
Amir Gholami, Kiseok Kwon, Bichen Wu, Zizheng Tai, Xiangyu Yue, Peter Jin,
  Sicheng Zhao, and Kurt Keutzer.
\newblock Squeezenext: Hardware-aware neural network design.
\newblock In \emph{Proceedings of the IEEE Conference on Computer Vision and
  Pattern Recognition Workshops}, pp.\  1638--1647, 2018.

\bibitem[He et~al.(2016)He, Zhang, Ren, and Sun]{he2016deep}
Kaiming He, Xiangyu Zhang, Shaoqing Ren, and Jian Sun.
\newblock Deep residual learning for image recognition.
\newblock In \emph{Proceedings of the IEEE conference on computer vision and
  pattern recognition}, pp.\  770--778, 2016.

\bibitem[Hinton et~al.(2012)Hinton, Deng, Yu, Dahl, Mohamed, Jaitly, Senior,
  Vanhoucke, Nguyen, Sainath, et~al.]{hinton2012deep}
Geoffrey Hinton, Li~Deng, Dong Yu, George~E Dahl, Abdel-rahman Mohamed, Navdeep
  Jaitly, Andrew Senior, Vincent Vanhoucke, Patrick Nguyen, Tara~N Sainath,
  et~al.
\newblock Deep neural networks for acoustic modeling in speech recognition: The
  shared views of four research groups.
\newblock \emph{IEEE Signal processing magazine}, 29\penalty0 (6):\penalty0
  82--97, 2012.

\bibitem[Hinton et~al.(2015)Hinton, Vinyals, and Dean]{hinton2015distilling}
Geoffrey Hinton, Oriol Vinyals, and Jeff Dean.
\newblock Distilling the knowledge in a neural network.
\newblock \emph{arXiv preprint arXiv:1503.02531}, 2015.

\bibitem[Hubara et~al.(2016)Hubara, Courbariaux, Soudry, El-Yaniv, and
  Bengio]{hubara2016binarized}
Itay Hubara, Matthieu Courbariaux, Daniel Soudry, Ran El-Yaniv, and Yoshua
  Bengio.
\newblock Binarized neural networks.
\newblock In \emph{Advances in neural information processing systems}, pp.\
  4107--4115, 2016.

\bibitem[Ioffe \& Szegedy(2015)Ioffe and Szegedy]{ioffe2015batch}
Sergey Ioffe and Christian Szegedy.
\newblock Batch normalization: Accelerating deep network training by reducing
  internal covariate shift.
\newblock \emph{arXiv preprint arXiv:1502.03167}, 2015.

\bibitem[Jouppi et~al.(2017)Jouppi, Young, Patil, Patterson, Agrawal, Bajwa,
  Bates, Bhatia, Boden, Borchers, et~al.]{jouppi2017datacenter}
Norman~P Jouppi, Cliff Young, Nishant Patil, David Patterson, Gaurav Agrawal,
  Raminder Bajwa, Sarah Bates, Suresh Bhatia, Nan Boden, Al~Borchers, et~al.
\newblock In-datacenter performance analysis of a tensor processing unit.
\newblock In \emph{Computer Architecture (ISCA), 2017 ACM/IEEE 44th Annual
  International Symposium on}, pp.\  1--12. IEEE, 2017.

\bibitem[Jung et~al.(2018)Jung, Son, Lee, Son, Kwak, Han, and
  Choi]{jung2018joint}
Sangil Jung, Changyong Son, Seohyung Lee, Jinwoo Son, Youngjun Kwak, Jae-Joon
  Han, and Changkyu Choi.
\newblock Joint training of low-precision neural network with quantization
  interval parameters.
\newblock \emph{arXiv preprint arXiv:1808.05779}, 2018.

\bibitem[Krizhevsky et~al.(2012)Krizhevsky, Sutskever, and
  Hinton]{krizhevsky2012imagenet}
Alex Krizhevsky, Ilya Sutskever, and Geoffrey~E Hinton.
\newblock Imagenet classification with deep convolutional neural networks.
\newblock In \emph{Advances in neural information processing systems}, pp.\
  1097--1105, 2012.

\bibitem[Li \& Liu(2016)Li and Liu]{li2016ternary}
F~Li and B~Liu.
\newblock Ternary weight networks.(2016).
\newblock \emph{arXiv preprint arXiv:1605.04711}, 2016.

\bibitem[Loshchilov \& Hutter(2016)Loshchilov and
  Hutter]{DBLP:journals/corr/LoshchilovH16a}
Ilya Loshchilov and Frank Hutter.
\newblock Sgdr: Stochastic gradient descent with warm restarts.
\newblock \emph{arXiv preprint arXiv:1608.03983}, 2016.

\bibitem[McKinstry et~al.(2018)McKinstry, Esser, Appuswamy, Bablani, Arthur,
  Yildiz, and Modha]{mckinstry2018discovering}
Jeffrey~L McKinstry, Steven~K Esser, Rathinakumar Appuswamy, Deepika Bablani,
  John~V Arthur, Izzet~B Yildiz, and Dharmendra~S Modha.
\newblock Discovering low-precision networks close to full-precision networks
  for efficient embedded inference.
\newblock \emph{arXiv preprint arXiv:1809.04191}, 2018.

\bibitem[Mishra \& Marr(2017)Mishra and Marr]{mishra2017apprentice}
Asit Mishra and Debbie Marr.
\newblock Apprentice: Using knowledge distillation techniques to improve
  low-precision network accuracy.
\newblock \emph{arXiv preprint arXiv:1711.05852}, 2017.

\bibitem[Polino et~al.(2018)Polino, Pascanu, and Alistarh]{polino2018model}
Antonio Polino, Razvan Pascanu, and Dan Alistarh.
\newblock Model compression via distillation and quantization.
\newblock \emph{arXiv preprint arXiv:1802.05668}, 2018.

\bibitem[Qiu et~al.(2016)Qiu, Wang, Yao, Guo, Li, Zhou, Yu, Tang, Xu, Song,
  et~al.]{qiu2016going}
Jiantao Qiu, Jie Wang, Song Yao, Kaiyuan Guo, Boxun Li, Erjin Zhou, Jincheng
  Yu, Tianqi Tang, Ningyi Xu, Sen Song, et~al.
\newblock Going deeper with embedded fpga platform for convolutional neural
  network.
\newblock In \emph{Proceedings of the 2016 ACM/SIGDA International Symposium on
  Field-Programmable Gate Arrays}, pp.\  26--35. ACM, 2016.

\bibitem[Rastegari et~al.(2016)Rastegari, Ordonez, Redmon, and
  Farhadi]{rastegari2016xnor}
Mohammad Rastegari, Vicente Ordonez, Joseph Redmon, and Ali Farhadi.
\newblock Xnor-net: Imagenet classification using binary convolutional neural
  networks.
\newblock In \emph{European Conference on Computer Vision}, pp.\  525--542.
  Springer, 2016.

\bibitem[Russakovsky et~al.(2015)Russakovsky, Deng, Su, Krause, Satheesh, Ma,
  Huang, Karpathy, Khosla, Bernstein, et~al.]{russakovsky2015imagenet}
Olga Russakovsky, Jia Deng, Hao Su, Jonathan Krause, Sanjeev Satheesh, Sean Ma,
  Zhiheng Huang, Andrej Karpathy, Aditya Khosla, Michael Bernstein, et~al.
\newblock Imagenet large scale visual recognition challenge.
\newblock \emph{International Journal of Computer Vision}, 115\penalty0
  (3):\penalty0 211--252, 2015.

\bibitem[Simonyan \& Zisserman(2014)Simonyan and Zisserman]{simonyan2014very}
Karen Simonyan and Andrew Zisserman.
\newblock Very deep convolutional networks for large-scale image recognition.
\newblock \emph{arXiv preprint arXiv:1409.1556}, 2014.

\bibitem[Sung et~al.(2015)Sung, Shin, and Hwang]{sung2015resiliency}
Wonyong Sung, Sungho Shin, and Kyuyeon Hwang.
\newblock Resiliency of deep neural networks under quantization.
\newblock \emph{arXiv preprint arXiv:1511.06488}, 2015.

\bibitem[Xu et~al.(2017)Xu, Gao, Yu, and Darrell]{xu2017end}
Huazhe Xu, Yang Gao, Fisher Yu, and Trevor Darrell.
\newblock End-to-end learning of driving models from large-scale video
  datasets.
\newblock In \emph{Proceedings of the IEEE conference on computer vision and
  pattern recognition}, pp.\  2174--2182, 2017.

\bibitem[You et~al.(2017)You, Gitman, and Ginsburg]{you2017large}
Yang You, Igor Gitman, and Boris Ginsburg.
\newblock Large batch training of convolutional networks.
\newblock \emph{arXiv preprint arXiv:1708.03888}, 2017.

\bibitem[Zhang et~al.(2018)Zhang, Yang, Ye, and Hua]{zhang2018lq}
Dongqing Zhang, Jiaolong Yang, Dongqiangzi Ye, and Gang Hua.
\newblock Lq-nets: Learned quantization for highly accurate and compact deep
  neural networks.
\newblock In \emph{Proceedings of the European Conference on Computer Vision
  (ECCV)}, pp.\  365--382, 2018.

\bibitem[Zhou et~al.(2016)Zhou, Wu, Ni, Zhou, Wen, and Zou]{zhou2016dorefa}
Shuchang Zhou, Yuxin Wu, Zekun Ni, Xinyu Zhou, He~Wen, and Yuheng Zou.
\newblock Dorefa-net: Training low bitwidth convolutional neural networks with
  low bitwidth gradients.
\newblock \emph{arXiv preprint arXiv:1606.06160}, 2016.

\bibitem[Zhu et~al.(2016)Zhu, Han, Mao, and Dally]{DBLP:journals/corr/ZhuHMD16}
Chenzhuo Zhu, Song Han, Huizi Mao, and William~J Dally.
\newblock Trained ternary quantization.
\newblock \emph{arXiv preprint arXiv:1612.01064}, 2016.

\end{thebibliography}
\bibliographystyle{iclr2020_conference}

\FloatBarrier

\appendix

\section{Step Size Gradient Scale Derivation}
\label{a:gradscale}

We compute our gradient scale value by first estimating $R$ (Equation \ref{eq:R}), starting with the simple heuristic that for a layer with $N_W$ weights
\begin{equation}
\sfrac{\lVert w \rVert}{ s } \approx \sqrt{N_W Q_P}.
\end{equation}
To develop this approximation, we first note that the expected value of an $l_2$-norm should grow with the square root of the number of elements normalized.
Next, we assume that where $Q_P=1$, step size should be approximately equal to average weight magnitude so as to split the weight distribution into zero and non zero values in a roughly balanced fashion.
Finally, we assume that for larger $Q_P$, step size should be roughly proportional to $\sqrt{\sfrac{1}{Q_P}}$, so that as the number of available quantized states increases, data between the clip points will be quantized more precisely, and the clip points themselves (equal to $sQ_N$ and $sQ_P$) will move further out to better encode outliers.

We also note that, in the expectation, $\lVert \nabla_w L \rVert$ and $\nabla_s L$ are of approximately the same order.
This can be shown by starting from the chain rule
\begin{equation}
	\nabla_s L = \sum_{i=1}^{N_W} \frac{\partial L}{\partial \hat{w_i}} \frac{\partial \hat{w}_i}{\partial s},
\end{equation}
then assuming $\sfrac{\partial \hat{w}_i}{\partial s}$ is reasonably close to 1 (see for example Figure \ref{fig:gradients}), and treating all $\sfrac{\partial L}{\partial \hat{w}_i}$ as uncorrelated zero-centered random variables, to compute the following expectation across weights:
\begin{equation}
	\text{E} \left[  \nabla_s L^2 \right] \approx
		N_W \times \text{E} \left[ \frac{\partial L}{\partial \hat{w}}^2 \right].
\end{equation}
By assuming $\sfrac{\partial \hat{w}}{\partial w}=1$ for most weights, we similarly approximate
\begin{equation}
	\text{E} \left[ \lVert \nabla_w L \rVert^2 \right] \approx
		N_W \times \text{E} \left[ \frac{\partial L}{\partial \hat{w}}^2 \right].
\end{equation}

Bringing all of this together, we can then estimate
\begin{equation}
R \approx \sqrt{ N_W Q_P }.
\end{equation}
Knowing this expected imbalance, we compute our gradient scale factor for weights by simply taking the inverse of $R$, so that $g$ is set to $\sfrac{1}{\sqrt{N_WQ_P}}$.

As most activation layers are preceded by batch normalization \citep{ioffe2015batch}, and assuming updates to the learned batch normalization scaling parameter is the primary driver of changes to pre-quantization activations, we can use a similar approach to the above to show that there is an imbalance between step size updates and update driven changes to activations that grows with the number of features in a layer, $N_F$ as well as $Q_P$.  Thus, for activation step size we set $g$ to $\sfrac{1}{\sqrt{N_FQ_P}}$.

\section{Implementation}

\newcommand{\algrule}[1][.2pt]{\par\vskip0.2\baselineskip\hrule height #1\par\vskip.2\baselineskip}

In this section we provide pseudocode to facilitate the implementation of LSQ.
We assume the use of automatic differentiation, as supported by a number of popular deep learning frameworks, where the desired operations for the training forward pass are coded, and the automatic differentiation engine computes the gradient through those operations in the backward pass.

Our approach requires two functions with non standard gradients, \textit{gradscale} (Function \ref{alg:gradscale}) and \textit{roundpass} (Function \ref{alg:roundpass}).
We implement the custom gradients by assuming a function called \textit{detach} that returns its input (unmodified) during the forward pass, and whose gradient during the backward pass is zero (thus detaching itself from the backward graph).
This function is used in the form:
\begin{equation}
y = detach(x_1 - x_2) + x_2,
\end{equation}
so that in the forward pass, $y=x_1$ (as the $x_2$ terms cancel out), while in the backward pass $\sfrac{\partial L}{\partial x_1} = 0$ (as detach blocks gradient propagation to $x_1$) and $\sfrac{\partial L}{\partial x_2} = \sfrac{\partial L}{\partial y}$.
We also assume a function $nfeatures$ that given an activation tensor, returns the number of features in that tensor, and $nweights$ that given a weight tensor, returns the number of weights in that tensor.
Finally, the above are used to implement a function called \textit{quantize}, which quantizes weights and activations prior to their use in each convolution or fully connected layer.

The pseudocode provided here is chosen for simplicity of implementation and broad applicability to many training frameworks, though more compute and memory efficient approaches are possible.
This example code assumes activations are unsigned, but could be modified to quantize signed activations.
%


\floatname{algorithm}{Function}
\setcounter{algorithm}{0}

\begin{algorithm}
   \caption{gradscale(x, scale):}
   \label{alg:gradscale}
\begin{algorithmic}
   \STATE \# x: Input tensor   
   \STATE \# scale: Scale gradient by this
   \STATE yOut = x
   \STATE yGrad = x $\times$ scale
   \STATE y = \textit{detach}(yOut - yGrad) + yGrad \# Return yOut in forward, pass gradient to yGrad in backward
   \STATE \textbf{return} y
\end{algorithmic}
\end{algorithm}

\begin{algorithm}
   \caption{roundpass(x):}
   \label{alg:roundpass}
\begin{algorithmic}
   \STATE \# x: Input tensor   
   \STATE yOut = \textit{round}(x) \# Round to nearest
   \STATE yGrad = x
   \STATE y = \textit{detach}(yOut - yGrad) + yGrad \# Return yOut in forward, pass gradient to yGrad in backward
   \STATE \textbf{return} y
\end{algorithmic}
\end{algorithm}

\begin{algorithm}
   \caption{quantize(v, s, p, isActivation):}
   \label{alg:quantize}
\begin{algorithmic}
   \STATE \# v: Input tensor
   \STATE \# s: Step size, a learnable parameter specific to weight or activation layer being quantized
   \STATE \# p: Quantization bits of precision
   \STATE \# isActivation: True if v is activation tensor, 
   \STATE \# \ \ \ \ \ \ \ \ \ \ \ \ \ \ \ \ \ \ \ \ \ \  False if v is weight tensor
   \STATE 
   \STATE \# Compute configuration values   
   \STATE {\bfseries if} isActivation:
   \STATE\hspace{\algorithmicindent} Qn = 0
   \STATE\hspace{\algorithmicindent} Qp = 2\^{}p - 1 
   \STATE\hspace{\algorithmicindent} gradScaleFactor = 1 / \textit{sqrt}(\textit{nfeatures}(v) $\times$ Qp)   
   \STATE {\bfseries else}: \# is weights
   \STATE\hspace{\algorithmicindent} Qn = -2\^{}(p-1)
   \STATE\hspace{\algorithmicindent} Qp = 2\^{}(p-1) - 1 
   \STATE\hspace{\algorithmicindent} gradScaleFactor = 1 / \textit{sqrt}(\textit{nweights}(v) $\times$ Qp)   
   \STATE
   \STATE \# Quantize
   \STATE s = \textit{gradscale}(s, gradScaleFactor)
   \STATE v = v / s
   \STATE v = \textit{clip}(v, Qn, Qp)
   \STATE vbar = \textit{roundpass}(v)
   \STATE vhat = vbar $\times$ s
   \STATE\textbf{return} vhat
\end{algorithmic}
\end{algorithm}

\end{document}